\documentclass[10pt,twocolumn,letterpaper]{article}

\usepackage{amsmath,amsfonts,bm}

\def\eqref#1{equation~\ref{#1}}

\def\1{\bm{1}}

\DeclareMathAlphabet{\mathsfit}{\encodingdefault}{\sfdefault}{m}{sl}
\SetMathAlphabet{\mathsfit}{bold}{\encodingdefault}{\sfdefault}{bx}{n}

\usepackage{wacv}              

\usepackage{graphicx}
\usepackage{amsmath}
\usepackage{amssymb}
\usepackage{booktabs}

\usepackage{amsfonts}       
\usepackage{xcolor}         
\usepackage{color, colortbl}

\usepackage{adjustbox}
\usepackage{multicol}
\usepackage{multirow}
\usepackage{units}
\usepackage{pifont}
\usepackage{dsfont}
\usepackage{subcaption}
\usepackage{wrapfig}
\usepackage{threeparttable}

\usepackage[pagebackref,breaklinks,colorlinks]{hyperref}

\usepackage[capitalize]{cleveref}
\crefname{section}{Sec.}{Secs.}
\Crefname{section}{Section}{Sections}
\Crefname{table}{Table}{Tables}
\crefname{table}{Tab.}{Tabs.}

\def\wacvPaperID{1961} 
\def\confName{WACV}
\def\confYear{2025}

\def\etal{\textit{et al. }}
\definecolor{Gray}{gray}{0.9}

\begin{document}

\title{Focusing on what to decode and what to train: \\SOV Decoding with Specific Target Guided DeNoising and \\Vision Language Advisor}

\author{
Junwen Chen \qquad  Yingcheng Wang \qquad  Keiji Yanai \\
\textit{Department of Informatics, The University of Electro-Communications, Tokyo, Japan}\\
{\tt\small \{chen-j, wang-y, yanai\}@mm.inf.uec.ac.jp}
}
\maketitle

\begin{abstract}
Recent transformer-based methods achieve notable gains in the Human-object Interaction Detection (HOID) task by leveraging the detection of DETR and the prior knowledge of Vision-Language Model (VLM).
However, these methods suffer from extended training times and complex optimization due to the entanglement of object detection and HOI recognition during the decoding process.
Especially, the query embeddings used to predict both labels and boxes suffer from ambiguous representations, and the gap between the prediction of HOI labels and verb labels is not considered.
To address these challenges, we introduce SOV-STG-VLA with three key components: Subject-Object-Verb (SOV) decoding, Specific Target Guided (STG) denoising, and a Vision-Language Advisor (VLA).
Our SOV decoders disentangle object detection and verb recognition with a novel interaction region representation.
The STG denoising strategy learns label embeddings with ground-truth information to guide the training and inference.
Our SOV-STG achieves a fast convergence speed and high accuracy and builds a foundation for the VLA to incorporate the prior knowledge of the VLM.
We introduce a vision advisor decoder to fuse both the interaction region information and the VLM's vision knowledge and a Verb-HOI prediction bridge to promote interaction representation learning.
Our VLA notably improves our SOV-STG and achieves SOTA performance with one-sixth of training epochs compared to recent SOTA.
Code and models are available at \url{https://github.com/cjw2021/SOV-STG-VLA}.
\end{abstract}

\begin{figure}[!t]
    \centering
    \resizebox{0.9\linewidth}{!}{
        \includegraphics[]{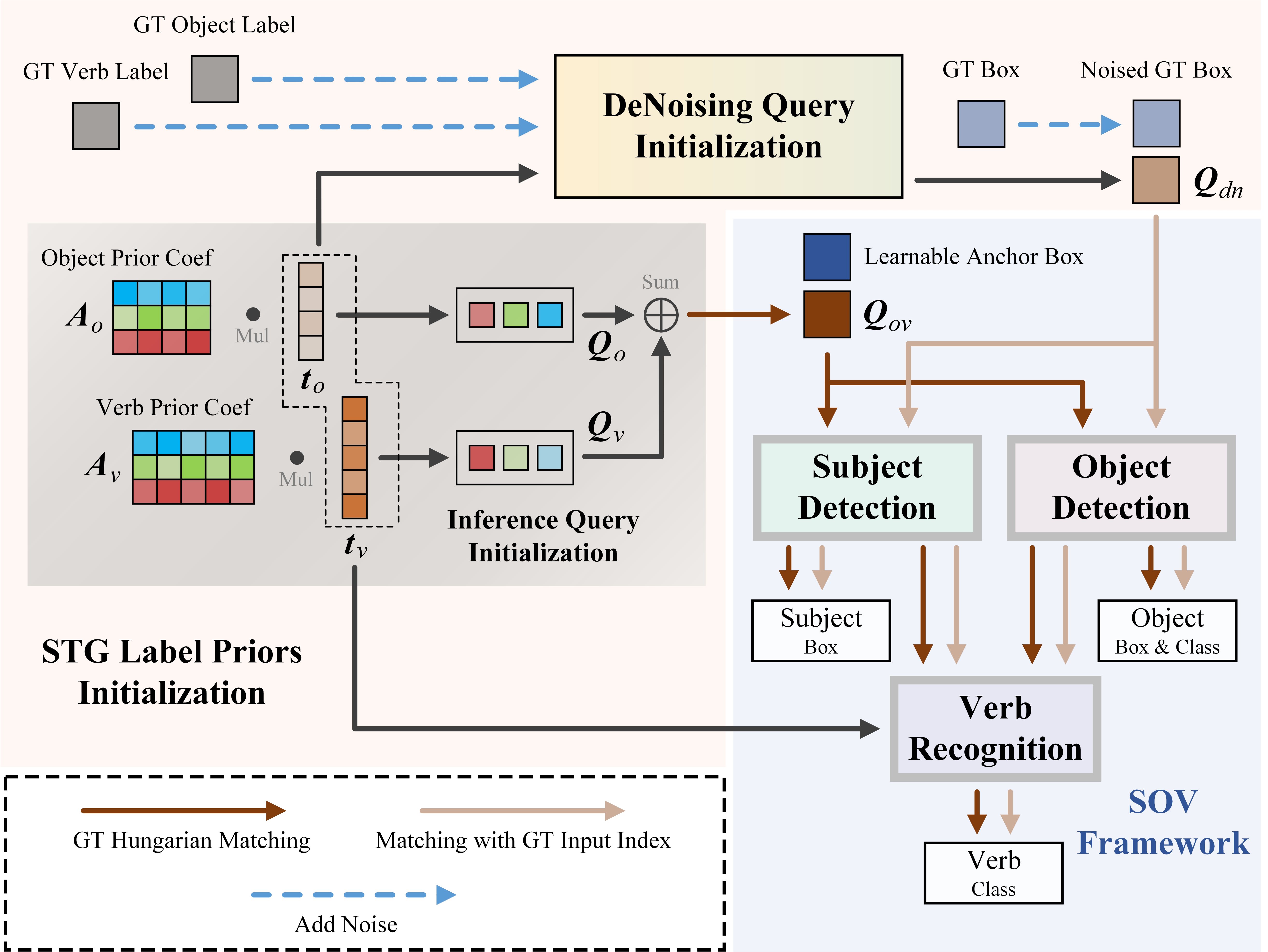}
    }
    \caption{
        \textbf{End-to-end training pipeline of our SOV-STG.} Our SOV framework splits the decoding process into three parts for each element of the HOI instance.
        Our STG training strategy efficiently transfers the ground-truth information to label embeddings through additional denoising queries.
    }
    \label{fig:SOV_STG_pipeline}
\end{figure}

\section{Introduction}
\label{sec:intro}

Recent Human-Object Interaction (HOI) detection studies are mainly built on the object detection framework.
The most widely used datasets, HICO-DET~\cite{chao2018learning} and V-COCO~\cite{gupta2015visual}, share the same object categories as the MS-COCO dataset~\cite{lin2014microsoft}.
Following the definition of the HOI instance $\{ B_s, (B_o, C_o), C_v \}$, which is a tuple of the subject (human) box $B_s$, the object box $B_o$ with class $C_o$, and the verb class $C_v$.
The HOI class $C_{\text{HOI}}$ is defined as the possible combination of the object class $C_o$ and the verb class $C_v$.
In the beginning, a multi-stream architecture built on top of a CNN-based object detector is commonly adopted in the two-stage methods~\cite{chao2018learning,gkioxari2018detecting,qi2018learning,gao2018ican}.
By introducing the human pose information~\cite{kim2020detecting,li2020detailed,zhong2021polysemy}, the language priors~\cite{gao2020drg,zhong2021polysemy}, or graph structure~\cite{gao2020drg,ulutan2020vsgnet,zhang2020spatio}, CNN-based methods achieve considerable accuracy.
On the other hand, CNN-based one-stage methods~\cite{liao2020ppdm,zhong2021glance,wang2020learning} leverage interaction points to detect possible interaction between the subject and object and achieve promising performance.

\begin{figure}[!t]
    \centering
    \resizebox{0.8\linewidth}{!}{
        \includegraphics[]{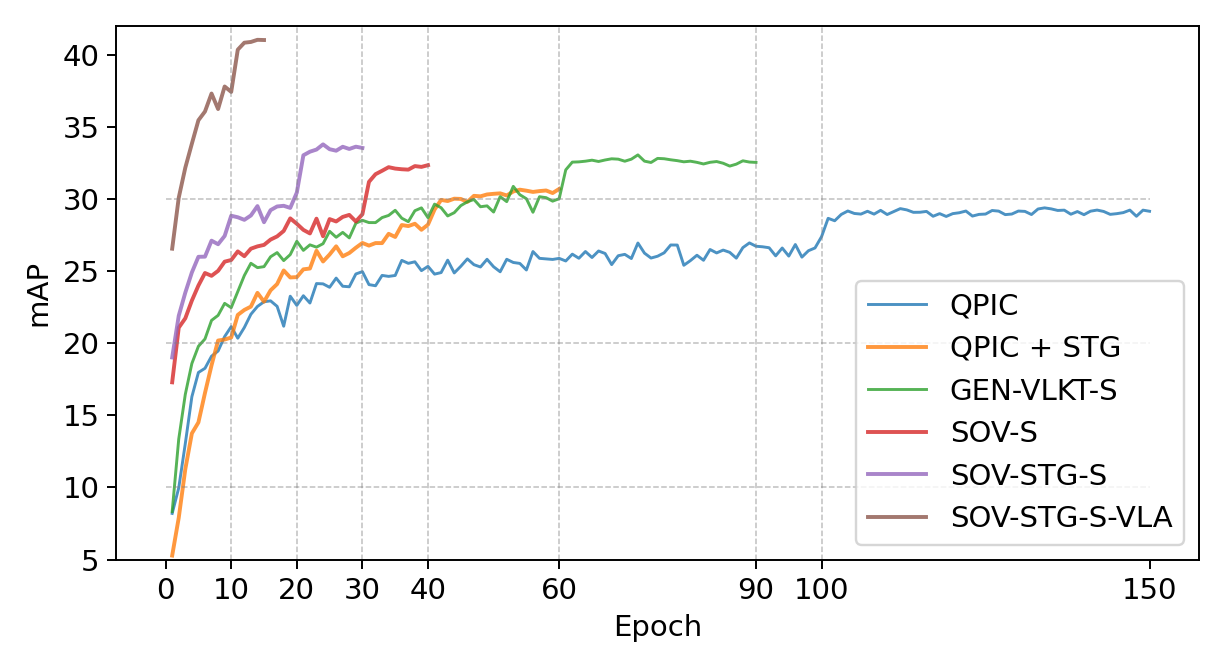}
    }
    \caption{
        Comparison of the training convergence curves of the state-of-the-art methods on the HICO-DET dataset.
    }
    \label{fig:ap_compare_sota}
\end{figure}

The attention mechanism of the transformer is more flexible than the CNN architecture in handling the relationships of features at different locations in the feature map and extracting global context information~\cite{dosovitskiy2021an}.
In the HOID task, transformer-based methods~\cite{tamura2021qpic,zou2021end,chen2021reformulating,kim2021hotr} show the advantage of the attention mechanism by adopting DETR~\cite{carion2020end}.
Without the matching process in one-stage and two-stage CNN-based methods, QPIC~\cite{tamura2021qpic} and HOITrans~\cite{zou2021end} adopt a compact encoder-decoder architecture to predict the HOI instances directly.
However, the compact architecture with a single decoder binds the feature of the subject and object localization and verb recognition together.
As a result, the finetuning for QPIC and HOITrans needs 150 and 250 epochs to converge, respectively.
Current one-stage methods~\cite{zhang2021mining,liao2022gen,yuan2022detecting,iftekhar2022look,zhou2022human,yuan2022rlip,kim2023relational,xie2023category} improve the single decoder design by disentangling the object localization and the verb recognition in a cascade manner.
Specifically, GEN-VLKT~\cite{liao2022gen} improves the cascade decoder design of CDN~\cite{zhang2021mining} by introducing two isolated queries of humans and objects in an instance decoder and fusing the human and object features in an interaction decoder.
However, the subject and object detection are still tangled in the instance decoder, and the spatial information is implicitly represented by the query embeddings.
Consequently, the training of GEN-VLKT is still hard, and it needs 90 epochs.
Recently, vision language model (VLM) based methods~\cite{ning2023hoiclip,mao2024clip4hoi,cao2024detecting} incorporates the pretrained VLM, CLIP~\cite{radford2021learning} or BLIP2~\cite{li2023blip} into the interaction decoding, which boosts the interaction representation learning.
However, the interaction decoder for the visual encoder of the VLM lacks the guidance of the spatial information of the interaction region to focus on the interaction region. Moreover, the prediction of HOI labels and verb labels with the HOI embeddings is not optimized.

To improve the training convergence and performance, our motivation can be summarized in three aspects: 1) how to focus on decoding specific targets, 2) how to guide the training with specific priors, and 3) how to align the pretrained VLM with the HOI and verb recognition.
For the first aspect, we revisit the decoding pipeline of the transformer-based method.
Recent one-stage methods~\cite{tamura2021qpic,zhang2021mining,liao2022gen} redirect the decoding target of the decoder pretrained from the object detection task, which leads to slow training convergence.
To this end, as shown in \cref{fig:SOV_STG_pipeline}, according to the definition of the HOI instance, we propose a new \textbf{S}ubject-\textbf{O}bject-\textbf{V}erb decoding framework, which splits the decoding process into three parts: subject detection, object detection, and verb recognition.
The spatial information (anchor boxes) and label information (label queries) are explicitly disentangled and fed into the decoders to guide the feature extraction.
focus on specific targets and share the training burden
In \cref{fig:ap_compare_sota}, we compare the training convergence with recent SOTA methods.
From the results, SOV takes advantage of the balanced decoding pipeline and the training converges faster.
The object detection part of SOV is the same as the pretrained detection model, which makes the training more stable and achieves a notable high accuracy at the early stage of the training.

For the second aspect, we focus on how to obtain specific label priors to initialize the label queries for HOI detection with an effective training strategy.
As shown in \cref{fig:SOV_STG_pipeline}, we introduce a novel \textbf{S}pcific \textbf{T}arget \textbf{G}uided (STG) denoising training strategy for HOI detection, which constructs a connection between the ground-truth label information and predefined label priors (embeddings) to guide the training.
Moreover, we leverage the verb label embeddings to guide the verb recognition in the verb recognition part to improve the verb representation learning capabilities.
In \cref{fig:ap_compare_sota}, we illustrate the training convergence of SOV and QPIC~\cite{tamura2021qpic} with STG, and the results show that our STG strategy effectively accelerates the training convergence before the learning rate drops and finally improves the performance.

With our SOV framework and STG training strategy, we introduce \textbf{V}ision \textbf{L}anguage \textbf{A}dvisor (VLA) to leverage the pretrained vision language model to further improve the interaction recognition and training convergence.
Our VLA consists of two parts: a vision advisor and a \textbf{V}erb-HOI (V-HOI) Bridge (language advisor).
VLA incorporates additional image-level information into the verb embeddings and considers the spatial information of the interaction region with our verb box representation.
V-HOI Bridge aims to fill the gap between the prediction of HOI labels and verb labels and aligns the pretrained VLM with the HOI recognition.
With the above advancements, our SOV-STG-VLA notably improves the performance and only needs 15 epochs to achieve the SOTA performance.


\begin{figure*}[!t]
    \centering
    \resizebox{0.9\linewidth}{!}{
        \includegraphics[]{img_arxiv/SOV-STG-VLA_WACV_down.jpg}
    }
    \caption{
        \textbf{The inference pipeline of SOV-STG-VLA.}
        SOV-STG consists of three parts: the STG label priors initialization, the subject and object detection, and the verb recognition.
        The label embeddings learned by our STG training strategy are used to initialize the label queries $\bm{Q}_{ov}$.
        The subject and object decoders update the learnable anchor boxes $B_s$ and $B_v$ to predict the subject and object, and the verb boxes $B_v$ are generated by our adaptive shifted MBR.
        Our SOV-STG-VLA is built on the SOV-STG framework.
        VLA enriches the expression of the verb embeddings $E_v$ by Vision Advisor with the global context information from the feature extractor and the pretrained VLM and the spatial information from the verb box.
        Then, V-HOI Bridge connects the prediction of HOI labels and verb labels.
    }
    \label{fig:overall_architecture}
\end{figure*}

\section{Related Work}

\noindent{\textbf{Predicting interactions with specific priors.}}\quad For one-stage transformer-based methods, how to extract the interaction information under a predefined representation of the interaction region is a key issue.
Recent studies~\cite{chen2023qahoi,ma2023fgahoi,Kim_2022_CVPR} attempt to leverage the deformable attention mechanism~\cite{zhu2020deformable} to guide the decoding by reference points.
QAHOI~\cite{chen2023qahoi} and FGAHOI~\cite{ma2023fgahoi} view the deformable transformer decoder's reference point as the HOI instance's anchor and use the anchor to guide the subject and object detection.
MSTR~\cite{Kim_2022_CVPR} proposes to use the subject, object, and context reference points to represent the HOI instance and predict the subject, object, and verb based on the reference points.
However, QAHOI, FGAHOI, and MSTR use x-y coordinates as the spatial priors to guide the decoding, the box size priors are not considered, and the query embeddings used to predict both labels and boxes suffer from ambiguous representations.
In contrast, our SOV explicitly defines the subject and object anchor boxes as the spatial priors and refines the anchor boxes layer by layer.
Moreover, for the verb recognition part of our SOV, we introduce a novel verb box to guide the verb feature extraction.

\noindent{\textbf{Effective learning with ground-truth guided.}}\quad For the object detection methods of the DETR family~\cite{carion2020end,zhu2020deformable,liu2022dabdetr}, DN-DETR~\cite{Li_2022_CVPR} shows that using the ground-truth information to guide the training can accelerate the training convergence and improve the performance.
In the HOID task, HQM~\cite{zhong2022towards} encodes the shifted ground-truth boxes as hard-positive queries to guide the training.
However, the ground-truth label information is not considered in HQM.
DOQ~\cite{qu2022distillation} introduces the oracle queries to implicitly encode the ground-truth boxes of human-object pairs and the object labels, and guide 
However, DOQ implicitly encodes the same number of oracle queries as the ground-truth with learnable weights and only uses the learned weights during training.
Without a complete and clear usage of ground-truth information, both HQM and DOQ still need 80 epochs to converge.
Different from DOQ and HQM, we introduce denoising queries to encode the ground-truth information and guide the training.
Moreover, our STG is used to learn the label priors for our model, and we intuitively use a ``select'' and a ``weighted sum'' approach to transfer the ground-truth label information to the denoising queries and inference queries, respectively.

\noindent{\textbf{VLM based HOID approaches.}}\quad Recent studies~\cite{jin2022overlooked,liao2022gen,ning2023hoiclip,mao2024clip4hoi} leverage the prior knowledge of the vision language model and achieve notable gains.
HOICLIP~\cite{ning2023hoiclip} and CLIP4HOI~\cite{mao2024clip4hoi} implement the pretrained CLIP model into the interaction decoding process.
UniHOI~\cite{cao2024detecting} introduces an HO Prompt Guideed Decoder to refine the HOI query embeddings with the visual features extracted by BLIP2~\cite{li2023blip}.
However, these three methods only predict the HOI labels or use the same decoder to predict the verb labels and HOI labels, which limits the representation of the HOI recognition prior knowledge of the VLM.
Instead of directly assigning the verb recognition task to the pretrained VLM, based on the spatial and refined label information provided by our SOV-STG, we introduce VLA to gradually incorporate the prior knowledge of the pretrained VLM into our framework and improve the interaction recognition.

\section{Methodology}

\cref{fig:overall_architecture} shows the inference pipeline of SOV-STG-VLA.
First, a feature extractor is used to extract the multi-scale global context features.
Then, the global features are fed into the object and subject decoder with learnable anchor boxes and label queries to predict pairs of subjects and objects.
The label queries are initialized by the label embeddings and learnable coefficient matrices as shown in \cref{fig:SOV_STG_pipeline}.
The STG denoising training strategy (In \cref{sec:split_label_embedding}) is used to learn the label embeddings with the ground-truth information.
The \textbf{S}ubject-\textbf{O}bject (S-O) attention module and \textbf{A}daptive \textbf{S}hifted \textbf{M}inimum \textbf{B}ounding \textbf{R}ectangle (ASMBR) used to fuse the subject and object embeddings and generate the verb box for the verb decoder are introduced in \cref{sec:split_decoder}.
Vision Advisor and V-HOI Bridge used to give additional guidance to our SOV-STG are introduced in \cref{sec:vision_advisor}.

\begin{figure}[!t]
    \centering
    \resizebox{0.8\linewidth}{!}{
        \includegraphics[]{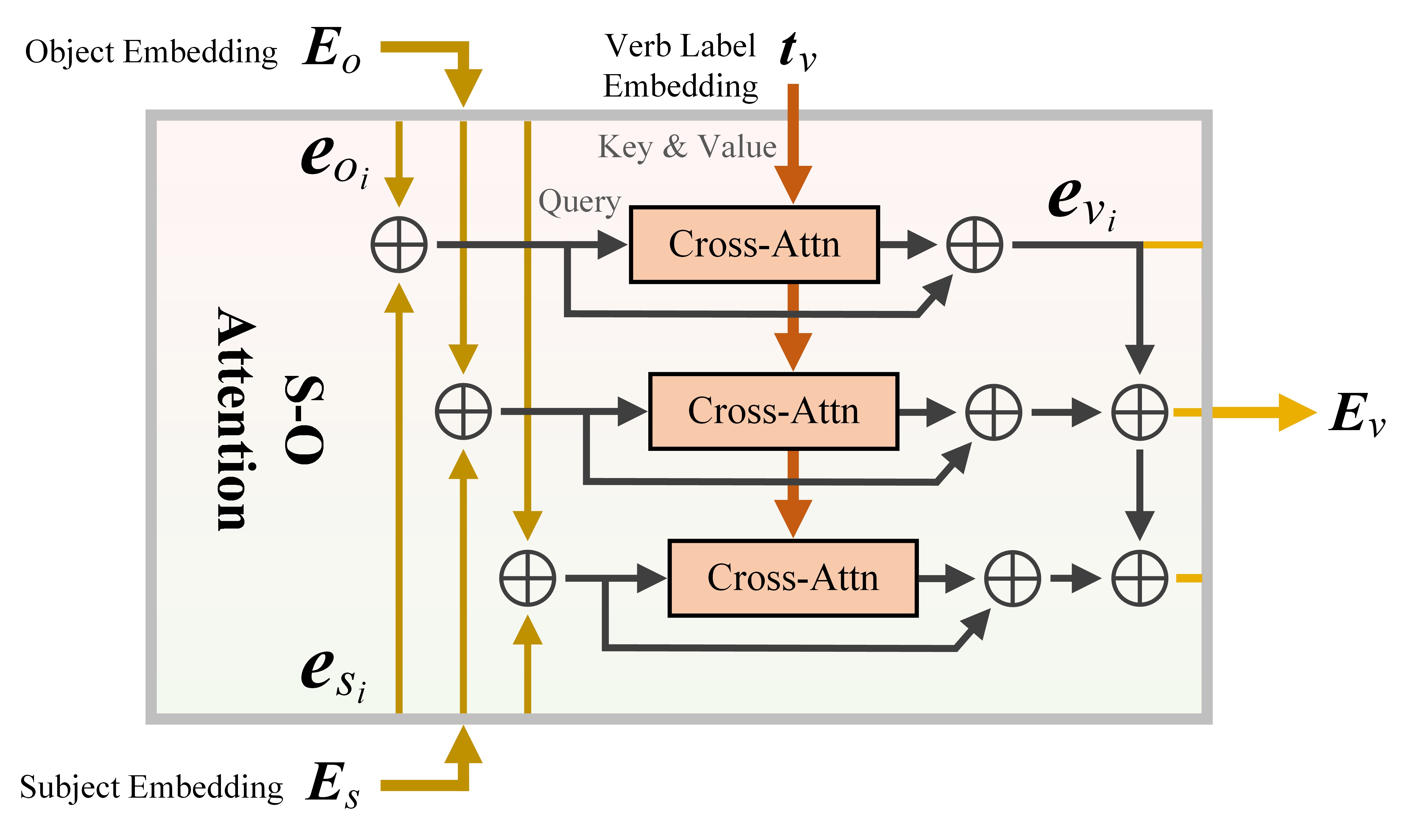}
    }
    \caption{
        The illustration of the S-O attention module.
    }
    \label{fig:so_attention}
\end{figure}

\subsection{SOV Decoding}
\label{sec:split_decoder}
To clarify the decoding target, the design of the split decoders is crucial for our framework.
Different from recent one-stage transformer-based methods~\cite{tamura2021qpic,zhang2021mining,liao2022gen}, which use a single decoder to detect objects and subjects,
we split the detection part into two decoders, the subject decoder, and the object decoder, and share the prediction burden of the verb decoder.
Moreover, we explore the design of the multi-branch feature fusion module and a new way to represent the interaction region for the verb decoder.

\noindent{\textbf{Subject and Object Detection.}}\quad First, we leverage a hierarchical backbone and deformable transformer encoder~\cite{zhu2020deformable} as the feature extractor to extract the multi-scale global features $\bm{f}_g$.
Then, we adopt an improved deformable transformer decoder proposed in the recent object detection method~\cite{liu2022dabdetr} as the object and subject decoder, which can process the label queries with the constraint of anchor boxes.
We clone the object decoder to initialize the subject decoder and alleviate the learning burden of the subject decoder.
As shown in \cref{fig:overall_architecture}, the subject and object decoder both use the label queries $\bm{Q}_{ov}\in\mathbb{R}^{N_q\times D}$ as the input queries, where $N_q$ is the number of queries and $D$ is the hidden dimension of the embeddings.
With the same query input, the pair of subject embeddings $\bm{E}_s\in\mathbb{R}^{N_l\times N_q\times D}$ and object embeddings $\bm{E}_o\in\mathbb{R}^{N_l\times N_q\times D}$ can share the same prior label information ($N_l$ indicates the number of the decoder layers).
The subject and object embeddings with corresponding learnable anchor boxes $\bm{B}_s$ and $\bm{B}_o$ are updated layer by layer during decoding.
Then, the object embeddings from the object decoder are used to predict the object classes.

\noindent{\textbf{Verb Decoder with S-O attention module.}}\quad As shown in \cref{fig:overall_architecture}, 
For the label queries of the verb decoder, we introduce S-O attention to fuse the subject and object embeddings in a multi-layer manner.
In \cref{fig:so_attention}, we illustrate the fusion process of our S-O attention module.
Given the subject embedding $\bm{e}_{s_i}$ and object embedding $\bm{e}_{o_i}$ from the $i$-th layer ($i>1$), first, we sum the subject and object embeddings to fuse an intermediate embedding $\bm{e}_{so_i}=(\bm{e}_{o_i}+\bm{e}_{s_i})/2$.
Then, to guide the verb recognition with our predefined label priors, the intermediate embeddings $\bm{e}_{so_i}$ are used to absorb the prior knowledge from the verb label embeddings $\bm{t}_{v}\in\mathbb{R}^{N_q\times D}$ learned by our STG training strategy (in \cref{sec:split_label_embedding}) through a cross-attention module.
Furthermore, we introduce a bottom-up path to amplify the information from the bottom to the top layer.
Finally, the verb embedding $\bm{e}_{v_i}$ after the bottom-up path can be defined as:
\begin{equation}
  \begin{split}
      \bm{e}_{v_i} = &  ( (\text{CrossAttn}(\bm{e}_{so_{i-1}}, \bm{t}_{v}) + \bm{e}_{so_{i-1}} ) \\
                     & + ( \text{CrossAttn}(\bm{e}_{so_{i}}, \bm{t}_{v}) + \bm{e}_{so_{i}} )  ) /2
  \end{split}
\end{equation}

\noindent{\textbf{Verb box represented by ASMBR}}\quad To constrain the verb feature extraction with positional information in the verb decoder, as shown in \cref{fig:overall_architecture}, we introduce a novel verb box, Adaptive Shifted Minimum Bounding Rectangle (ASMBR) as the representation of the interaction region.
The verb box is directly initialized from the last layer of subject and object boxes from the subject and object decoder.
To balance the attention between the subject and object, we shift the center of the MBR to the center of the subject and object boxes.
Considering the boxes will overlap with each other, we shrink the width and height of the MBR according to the spatial relationship between the two boxes.
With the shift and adapt operation, the verb box can constrain the interaction region for sampling points of the deformable attention and extract interaction information from specific subject and object pairs.
Finally, given the last layer subject box $\bm{B}_s=(x_s, y_s, w_s, h_s)$ and object box $\bm{B}_o=(x_o, y_o, w_o, h_o)$, where $(x,y)$ indicates the box center, the verb box is defined as:
\begin{equation}
    \bm{B}_{v} = \bigg(\frac{x_s + x_o}{2}, \frac{y_s + y_o}{2}, w_{v}, h_{v}\bigg)
\end{equation}
\begin{equation}
    w_{v} = \frac{w_s + w_o}{2}+|x_s-x_o|, h_{v} = \frac{h_s + h_o}{2}+|y_s-y_o|
\end{equation}

\subsection{Specific Target Guided DeNoising Training}
\label{sec:split_label_embedding}
For our SOV framework, we generate inference and denoising label queries during training.
Since the two kinds of queries are generated from the specific target priors and learned during the denoising training, we call our training strategy as \textbf{S}pecific \textbf{T}arget \textbf{G}uided (STG) denoising.
\noindent{\textbf{Label-specific Priors}}\quad To explicitly equip the prior label knowledge into the decoders and disentangle the training and decoding target,
as shown in \cref{fig:SOV_STG_pipeline}, two kinds of learnable label embeddings are used to initialize the query embeddings.
Specifically, we define the object label embeddings $\bm{t}_o\in \mathbb{R}^{C_o\times D}$ as the object label priors, which consist of $C_o$ vectors with $D$ dimensions, where $C_o$ is the number of object classes.
Similarly, the verb label embeddings $\bm{t}_v\in \mathbb{R}^{C_v\times D}$ are defined as the verb label priors.
With the object label and verb label priors, we first initialize the query embeddings of object label $\bm{q}_o\in\mathbb{R}^{N_q\times D}$ and verb label $\bm{q}_v\in\mathbb{R}^{N_q\times D}$ by linear combining the object label and verb label embeddings with two learnable coefficient matrices $\bm{A}_o\in\mathbb{R}^{N_q\times C_o}$ and $\bm{A}_v\in\mathbb{R}^{N_q\times C_v}$, respectively.
Then, we add the object and verb label embeddings to obtain the inference query embeddings $\bm{q}_{ov}\in \mathbb{R}^{N_q \times D}$.
The initialization of $\bm{q}_o$, $\bm{q}_v$, and $\bm{q}_{ov}$ is defined as follows:
\begin{equation}
    \bm{q}_o = \bm{A}_o \bm{t}_o, \quad \bm{q}_v = \bm{A}_v \bm{t}_v
\end{equation}
\begin{equation}
    \bm{q}_{ov} = \bm{q}_o + \bm{q}_v
\end{equation}
Different from DN-DETR~\cite{Li_2022_CVPR} and DOQ~\cite{qu2022distillation}, which learn an encoding weight to generate queries only used in training,
we use the label embeddings both in the denoising and inference parts and enable the inference part to obtain the input query with label-specific information from the beginning.

\begin{figure}[!t]
    \centering
    \resizebox{0.9\linewidth}{!}{
        \includegraphics[]{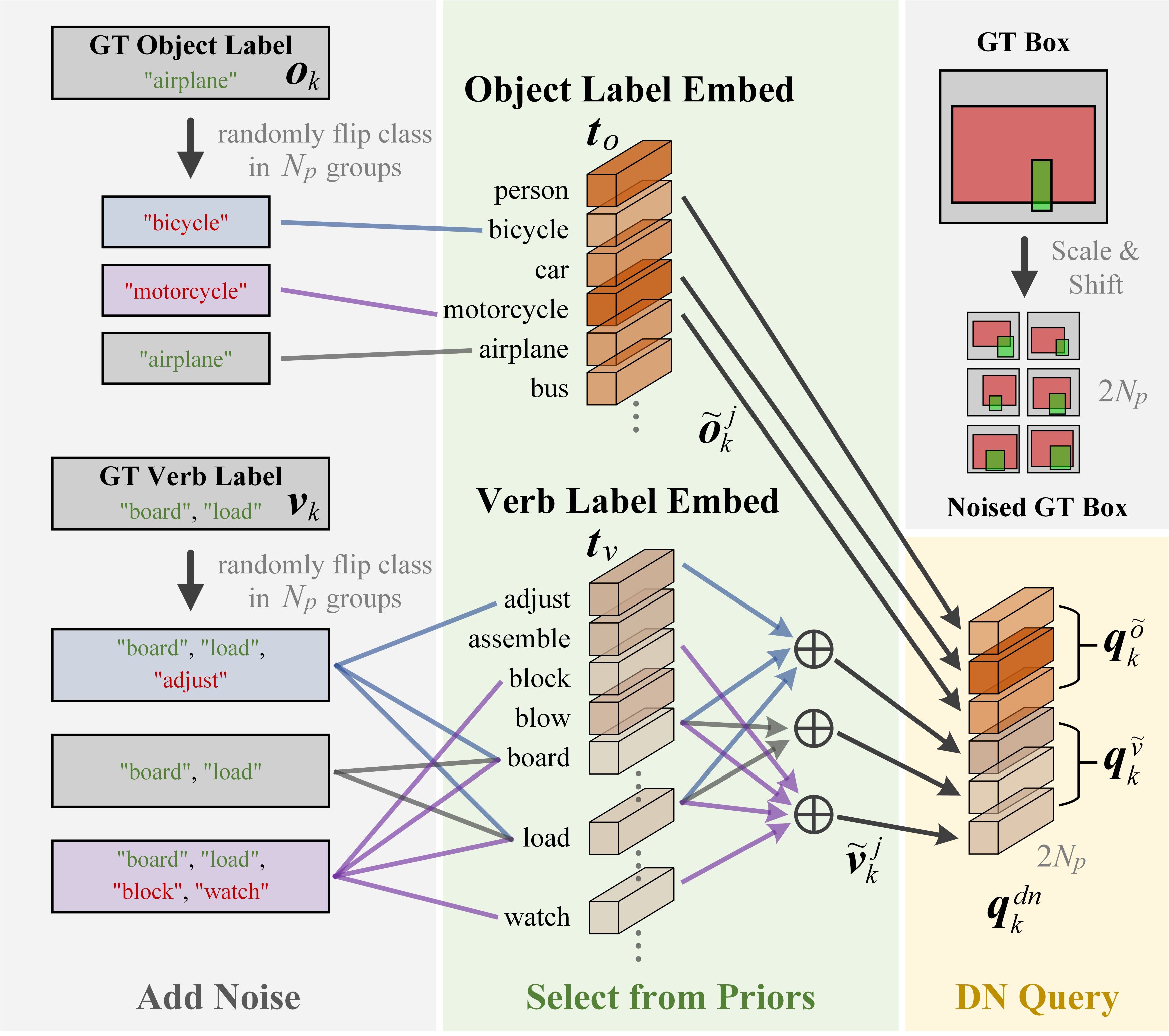}
    }
    \caption{
        \textbf{Illustration of adding noise to a ground-truth HOI instance.}
        The initialization consists of two parts, the object label and the verb label DN queries initialization.
        The final DN query embeddings $\bm{q}^{dn}_{k}$ are concatenated with the object label DN queries $\bm{q}^{\tilde{o}}_{k}$ and the verb label DN queries $\bm{q}^{\tilde{v}}_{k}$.
    }
    \label{fig:dn_query_init}
\end{figure}

\noindent{\textbf{Learning Priors with DeNoising Training}}\quad In \cref{fig:dn_query_init}, we show the initialization of the DN (DeNoising) query embeddings and visualize the process of adding noise to a ground-truth HOI instance.
Given the ground-truth object label set $\bm{O}_{gt}=\{\bm{o}_i\}_{i=1}^K$ and verb label set $\bm{V}_{gt}=\{\bm{v}_i\}_{i=1}^K$ of an image, where $\bm{o}_i$ and $\bm{v}_i$ are the labels of the object and verb classes, $K$ is the number of ground-truth HOI instances, we generate $N_p$ groups of noised labels for each of the ground-truth HOI instances.
For the $k$-th ground-truth HOI instance, the noised object labels are obtained by randomly flipping the ground-truth index of the object label $\bm{o}_k$ to another object class index.
Because the verb label $\bm{v}_k$ consists of co-occurrence ground-truth classes, to keep the co-occurrence ground-truth indices appearing in the noised verb labels, we randomly flip the other indices of the ground-truth verb label to generate the noised verb labels.
Two flipping rate hyper-parameters $\eta_o\in(0,1)$ and $\eta_v\in(0,1)$ are used to control the percentage of the noised object labels and verb labels, respectively.
Besides, a verb class flipping rate hyper-parameter $\lambda_v\in(0,1)$ is used to control the class-wise flipping rate in the verb labels.
Next, we introduce a "select" approach to "encode" the noised labels to DN query embeddings.
Specifically, we directly compose the object DN query embeddings $\bm{q}^{\tilde{o}}_k\in\mathbb{R}^{N_p\times D}$ by selecting class-specific vectors $\{\tilde{\bm{o}}^j_k\}_{j=1}^{N_p}$ from the object label embeddings $\bm{t}_o$ according to the indices of the noised object labels.
For encoding of the noised verb labels, we select and sum the class-specific vectors to construct multi-class vectors $\{\tilde{\bm{v}}^j_k\}_{j=1}^{N_p}$, and compose the verb DN query embeddings $\bm{q}^{\tilde{v}}_k\in\mathbb{R}^{N_p\times D}$.
Finally, we concatenate the object DN query embeddings $\bm{q}^{\tilde{o}}_k$ and verb DN query embeddings $\bm{q}^{\tilde{v}}_k$ to form the DN query embeddings $\bm{q}^{dn}_k\in\mathbb{R}^{2N_p\times D}$ for the denoising training.
Since the specific target priors learned by the denoising training are also used to guide the inference during end-to-end training, our STG can accelerate the training convergence and improve the inference performance at the same time.
In addition, motivated by the box denoising strategy of DN-DETR~\cite{Li_2022_CVPR}, we scale and shift pairs of ground-truth subject and object boxes to generate $2N_p$ groups of noised anchor boxes for corresponding DN query embeddings.

\begin{figure*}[!t]
    \begin{minipage}{0.49\linewidth}
        \centering
        \resizebox{0.63\linewidth}{!}{
            \includegraphics[]{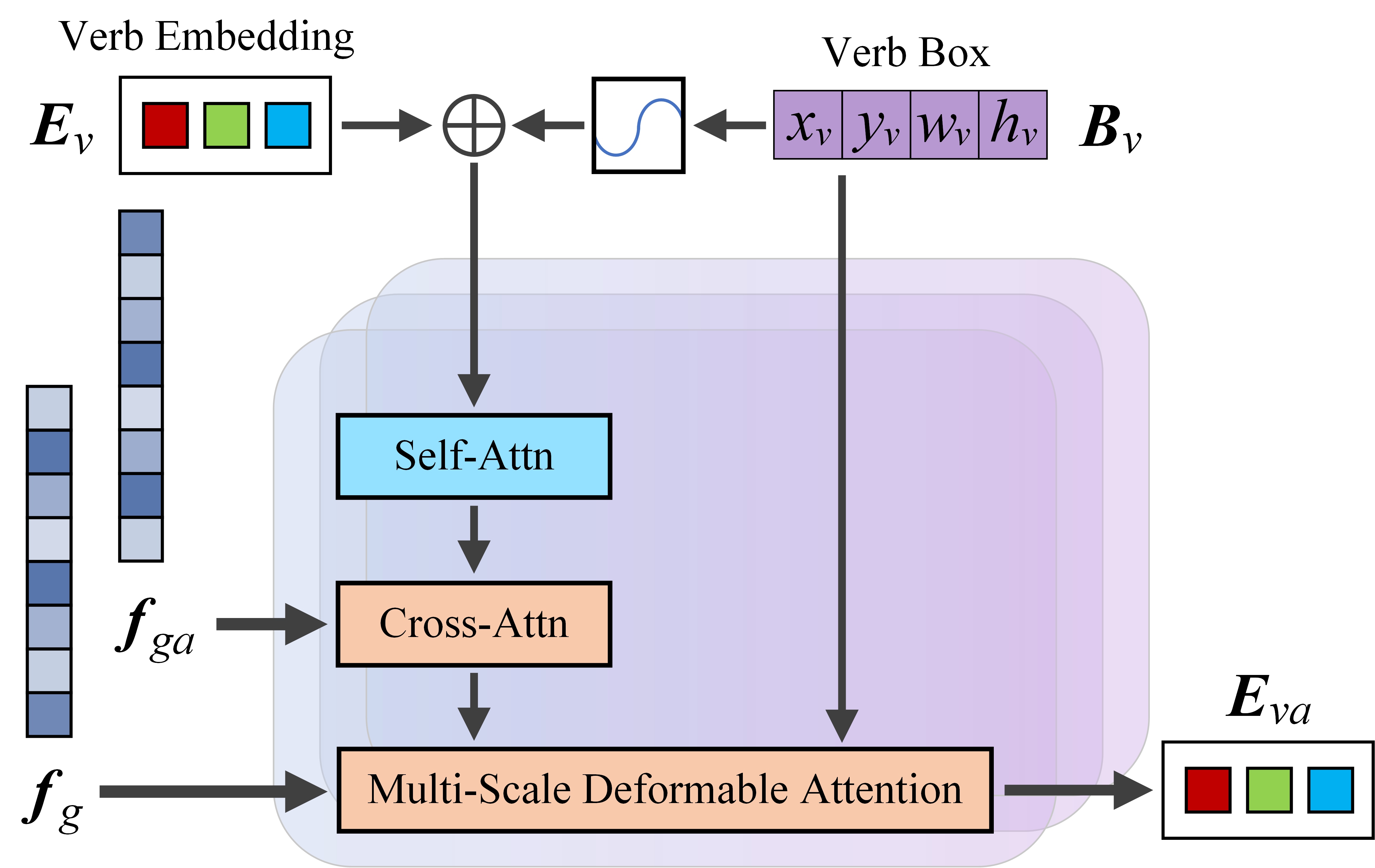}
        }
        \caption{
            The illustration of the vision advisor decoder.
        }
        \label{fig:vison_advisor_decoder}
    \end{minipage}
    \hfill
    \begin{minipage}{0.49\linewidth}
        \centering
        \resizebox{0.935\linewidth}{!}{
            \includegraphics[]{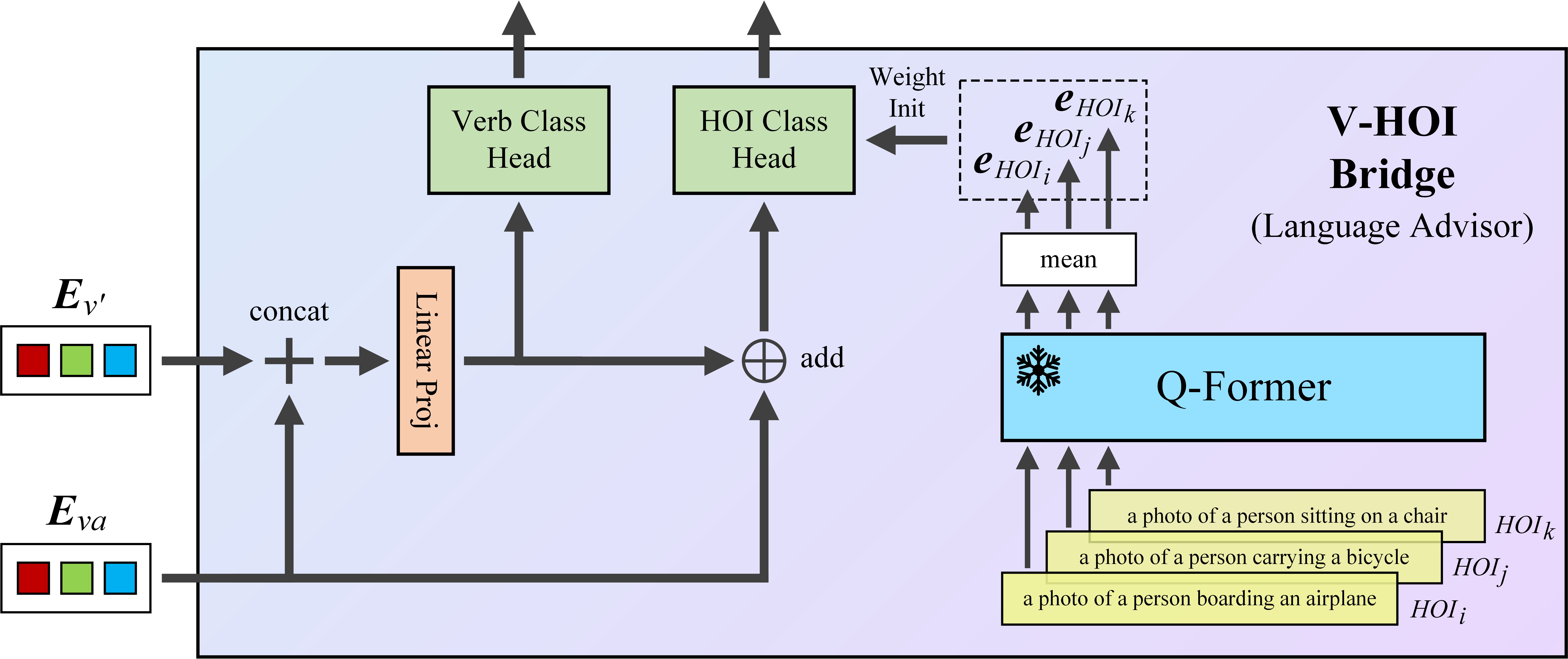}
        }
        \caption{
            The illustration of V-HOI Bridge.
        }
        \label{fig:v-hoi-bridge}
    \end{minipage}
\end{figure*}

\subsection{Vision Language Advisor}
\label{sec:vision_advisor}
Our Vision Language Advisor (VLA) is designed to release the burden of aligning the pretrained VLM knowledge with the verb recognition part of SOV-STG and fill the gap between the verb and HOI label prediction.

\noindent{\textbf{Vision Advisor.}}\quad We leverage the visual encoder and Q-Former of BLIP2~\cite{li2023blip} to extract the image-level features $\bm{f}_{ga}\in\mathbb{R}^{N_{ga}\times {D_a}}$.
We introduce the vision advisor decoder to incorporate the global visual features $\bm{f}_g$ and $\bm{f}_{ga}$ from our SOV model and the vision advisor into our verb embeddings $\bm{E}_v$ while considering the spatial information of interaction region.
To better connect the visual knowledge to each HOI query embedding, as shown in \cref{fig:vison_advisor_decoder}, we design a cascade attention mechanism for our vision advisor decoder.
Different from recent VLM-based HOID methods~\cite{ning2023hoiclip,cao2024detecting,mao2024clip4hoi} using the self-attention without the positional embedding,
we convert the verb box to a positional embedding through a positional encoding method~\cite{liu2022dabdetr} for the self-attention module.
Then, the same as the original transformer decoder, a cross-attention is used to extract the global context information from the visual encoder of the VLM.
In addition to aligning the prior knowledge with the verb recognition part, we add a multi-scale deformable attention module~\cite{zhu2020deformable} after the cross-attention to focus on the region of the verb box.

For each layer of the vision advisor decoder, the refined verb embeddings $e_{va}$ can be calculated as follows:
\begin{equation}
    \bm{e}_{va} = \text{MSACrossAttn}(\bm{e}_{t}, \bm{f}_g, \bm{B}_v)
\end{equation}
\begin{equation}
    \bm{e}_{t} = \text{MHACrossAttn}(\text{SelfAttn}(\bm{e}_{v}, \text{PE}(\bm{B}_v)), \bm{f}_{ga})
\end{equation}
where $\text{MSACrossAttn}$, $\text{MHACrossAttn}$, and $\text{PE}$ are multi-scale, multi-head cross-attention mechanisms, and positional encoding, respectively.
With the MHACrossAttn, the vision advisor is able to focus on the region of the verb box, which is related to each HOI query embedding.

\noindent{\textbf{Verb-HOI Bridge.}}\quad For our Verb-HOI Bridge (also language advisor) in \cref{fig:v-hoi-bridge}, we predict the HOI classes in a two-step manner to fill the gap between the verb recognition and HOI recognition.
In the first step, we concatenate $\bm{E}_{v'}$ after our verb decoder and $\bm{E}_{va}$ from the vision advisor decoder and feed into a linear projection layer to form $E_{vt}$ and predict the verb classes with a linear verb prediction head in original SOV-STG framework.
As the first step follows the original training pipeline of SOV-STG, the vision advisor can align VLM's visual knowledge in a straight feed and return style.
In the second step, we add $\bm{E}_{vt}$ to $\bm{E}_{va}$ and predict the HOI classes with the HOI prediction head.
Similar to previous VLM-based HOID methods~\cite{jin2022overlooked,liao2022gen,ning2023hoiclip}, we generate text embeddings to initialize the weights of the HOI prediction head.
However, different from previous works that use CLIP text encoder~\cite{liao2022gen,ning2023hoiclip} to encode the HOI text prompt into a single vector, we keep the same instruction as the visual feature extraction, we use Q-Former to encode the HOI text prompt into a set of vectors and use the average of the set as the weight of the HOI classifier.
Since BLIP2 introduced Q-Former as the bridge between the visual and language information and complete the relation between the vision advisor and the language advisor, thus, we call this part as V-HOI Bridge.

\subsection{Training and Inference}
\label{sec:training_inference}
As shown in \cref{fig:SOV_STG_pipeline}, our proposed method SOV-STG is trained in an end-to-end manner.
For the inference queries $\bm{Q}_{ov}$, the Hungarian algorithm~\cite{kuhn1955hungarian} is used to match the ground-truth HOI instances with the predicted HOI instances, and the matching cost and the training loss of predicted HOI instances follow the previous deformable transformer based method~\cite{chen2023qahoi}.
For the DN queries $\bm{Q}_{dn}$, the ground-truth indices used in query initialization are used to match the predicted HOI instances, and the loss function is the same as the inference queries.
As the label embeddings used in the denoising training part are also the specific target priors of the inference part, SOV-STG uses all of the parameters in training and inference.

\begin{table*}[!t]
  \begin{minipage}{0.64\linewidth}
      \centering
      \resizebox{0.95\linewidth}{!}{
          \begin{tabular}{@{}ccccccccc@{}}
              \hline
              \multicolumn{1}{c|}{}                                       & \multicolumn{1}{c|}{}                & \multicolumn{1}{c|}{}                & \multicolumn{3}{c|}{Default} & \multicolumn{3}{c}{Known Object} \\
              \multicolumn{1}{c|}{Method}                                 & \multicolumn{1}{c|}{Epoch}           & \multicolumn{1}{c|}{Backbone}        & \textit{Full}                & \textit{Rare}    & \multicolumn{1}{c|}{\textit{Non-Rare}} & \textit{Full}    & \textit{Rare}    & \textit{Non-Rare} \\ \hline \hline
              \multicolumn{1}{l}{\textbf{Two-stage}} \\ \hline   
              \multicolumn{1}{l|}{CATN~\cite{dong2022category}}           & \multicolumn{1}{c|}{12}              & \multicolumn{1}{c|}{ResNet-50}       & 31.86           & 25.15            & \multicolumn{1}{c|}{33.84}             & 34.44            & 27.69            & 36.45    \\
              \multicolumn{1}{l|}{STIP~\cite{zhang2022exploring}}         & \multicolumn{1}{c|}{30}              & \multicolumn{1}{c|}{ResNet-50}       & 32.22           & 28.15            & \multicolumn{1}{c|}{33.43}             & 35.29            & 31.43            & 36.45    \\
              \multicolumn{1}{l|}{UPT~\cite{Zhang_2022_CVPR}}             & \multicolumn{1}{c|}{20}              & \multicolumn{1}{c|}{ResNet-101-DC5}  & 32.62           & 28.62            & \multicolumn{1}{c|}{33.81}             & 36.08            & 31.41            & 37.47    \\ 
              \multicolumn{1}{l|}{Liu~\etal~\cite{Liu_2022_CVPR}}         & \multicolumn{1}{c|}{129}             & \multicolumn{1}{c|}{ResNet-50}       & 33.51           & 30.30            & \multicolumn{1}{c|}{34.46}             & 36.28            & 33.16            & 37.21    \\ 
              \multicolumn{1}{l|}{ViPLO~\cite{park2023viplo}}             & \multicolumn{1}{c|}{8}               & \multicolumn{1}{c|}{ViT-B/32}        & 34.95           & 33.83            & \multicolumn{1}{c|}{35.28}             & 38.15            & 36.77            & 38.56    \\ \hline \hline
              \multicolumn{1}{l}{\textbf{One-stage}} \\ \hline
              \multicolumn{1}{l|}{QAHOI~\cite{chen2023qahoi}}                 & \multicolumn{1}{c|}{150}             & \multicolumn{1}{c|}{ResNet-50}       & 26.18           & 18.06            & \multicolumn{1}{c|}{28.61}             & -                & -                & -                 \\ 
              \multicolumn{1}{l|}{QPIC~\cite{tamura2021qpic}}             & \multicolumn{1}{c|}{150}             & \multicolumn{1}{c|}{ResNet-50}       & 29.07           & 21.85            & \multicolumn{1}{c|}{31.23}             & 31.68            & 24.14            & 33.93             \\ 
              \multicolumn{1}{l|}{MSTR~\cite{Kim_2022_CVPR}}              & \multicolumn{1}{c|}{50}              & \multicolumn{1}{c|}{ResNet-50}       & 31.17           & 25.31            & \multicolumn{1}{c|}{32.92}             & 34.02            & 28.83            & 35.57             \\
              \multicolumn{1}{l|}{CDN-L~\cite{zhang2021mining}}           & \multicolumn{1}{c|}{100}             & \multicolumn{1}{c|}{ResNet-101}      & 32.07           & 27.19            & \multicolumn{1}{c|}{33.53}             & 34.79            & 29.48            & 36.38             \\
              \multicolumn{1}{l|}{HQM (CDN-S)~\cite{zhong2022towards}}    & \multicolumn{1}{c|}{80}              & \multicolumn{1}{c|}{ResNet-50}       & 32.47           & 28.15            & \multicolumn{1}{c|}{33.76}             & 35.17            & 30.73            & 36.50             \\
              \multicolumn{1}{l|}{RLIP-ParSe~\cite{yuan2022rlip}}         & \multicolumn{1}{c|}{90}              & \multicolumn{1}{c|}{ResNet-50}       & 32.84           & 34.63            & \multicolumn{1}{c|}{26.85}             & -                & -                & -                 \\
              \multicolumn{1}{l|}{DOQ (CDN-S)~\cite{qu2022distillation}}  & \multicolumn{1}{c|}{80}              & \multicolumn{1}{c|}{ResNet-50}       & 33.28           & 29.19            & \multicolumn{1}{c|}{34.50}             & -                & -                & -                 \\
              \multicolumn{1}{l|}{GEN-VLKT-S~\cite{liao2022gen}}          & \multicolumn{1}{c|}{90}              & \multicolumn{1}{c|}{ResNet-50}       & 33.75           & 29.25            & \multicolumn{1}{c|}{35.10}             & 36.78            & 32.75            & 37.99             \\ 
              \multicolumn{1}{l|}{HOICLIP~\cite{ning2023hoiclip}}         & \multicolumn{1}{c|}{90}              & \multicolumn{1}{c|}{ResNet-50}       & 34.69           & 31.12            & \multicolumn{1}{c|}{35.74}             & 37.61            & 34.47            & 38.54             \\ 
              \multicolumn{1}{l|}{GEN-VLKT-M~\cite{liao2022gen}}          & \multicolumn{1}{c|}{90}              & \multicolumn{1}{c|}{ResNet-101}      & 34.78           & 31.50            & \multicolumn{1}{c|}{35.77}             & 38.07            & 34.94            & 39.01             \\
              \multicolumn{1}{l|}{GEN-VLKT-L~\cite{liao2022gen}}          & \multicolumn{1}{c|}{90}              & \multicolumn{1}{c|}{ResNet-101}      & 34.95           & 31.18            & \multicolumn{1}{c|}{36.08}             & 38.22            & 34.36            & 39.37    \\ 
              \multicolumn{1}{l|}{CLIP4HOI~\cite{mao2024clip4hoi}}        & \multicolumn{1}{c|}{100}             & \multicolumn{1}{c|}{ResNet-50}       & 35.33           & 33.95            & \multicolumn{1}{c|}{35.74}             & 37.19            & 35.27            & 37.77             \\ 
              \multicolumn{1}{l|}{CQL(+GEN-VLKT-S)~\cite{xie2023category}} & \multicolumn{1}{c|}{90}             & \multicolumn{1}{c|}{ResNet-50}       & 35.36           & 32.97            & \multicolumn{1}{c|}{36.07}             & 38.43            & 34.85   & 39.50             \\ 
              \multicolumn{1}{l|}{UniHOI-S~\cite{cao2024detecting}}       & \multicolumn{1}{c|}{90}              & \multicolumn{1}{c|}{ResNet-50}      & 40.06           & \textbf{39.91}   & \multicolumn{1}{c|}{40.11}             & 42.20            & 42.60   & 42.08            \\ 
              \rowcolor{Gray}
              \multicolumn{1}{l|}{QAHOI-Swin-L~\cite{chen2023qahoi}}          & \multicolumn{1}{c|}{150}             & \multicolumn{1}{c|}{Swin-Large-22K}  & 35.78           & 29.80            & \multicolumn{1}{c|}{37.56}             & 37.59            & 31.36            & 39.36             \\
              \rowcolor{Gray}
              \multicolumn{1}{l|}{FGAHOI-Swin-L~\cite{ma2023fgahoi}}      & \multicolumn{1}{c|}{190}             & \multicolumn{1}{c|}{Swin-Large-22K}  & 37.18           & 30.71            & \multicolumn{1}{c|}{39.11}             & 38.93            & 31.93            & 41.02             \\
              \rowcolor{Gray}
              \multicolumn{1}{l|}{DiffHOI-Swin-L~\cite{yang2023boosting}} & \multicolumn{1}{c|}{90}              & \multicolumn{1}{c|}{Swin-Large-22K}  & 41.50           & 39.96            & \multicolumn{1}{c|}{41.96}             & 43.62            & 41.41            & 44.28             \\
              \rowcolor{Gray}
              \multicolumn{1}{l|}{PViC-Swin-L~\cite{zhang2023pvic}}       & \multicolumn{1}{c|}{30}              & \multicolumn{1}{c|}{Swin-Large}      & 44.32           & \textbf{44.61}   & \multicolumn{1}{c|}{44.24}             & 47.81            & \textbf{48.38}   & 47.64             \\
              \rowcolor{Gray}
              \multicolumn{1}{l|}{RLIPv2-ParSeDA-Swin-L~\cite{yuan2023rlipv2}} & \multicolumn{1}{c|}{20}         & \multicolumn{1}{c|}{Swin-Large}      & 45.09           & 43.23            & \multicolumn{1}{c|}{45.64}             & -                & -                & -             \\ \cline{1-9}
              \multicolumn{1}{l|}{\textbf{SOV-STG-S}}                     & \multicolumn{1}{c|}{30}              & \multicolumn{1}{c|}{ResNet-50}       & 33.80	          & 29.28            & \multicolumn{1}{c|}{35.15}             & 36.22            & 30.99            & 37.78             \\
              \multicolumn{1}{l|}{\textbf{SOV-STG-M}}                     & \multicolumn{1}{c|}{30}              & \multicolumn{1}{c|}{ResNet-101}      & 34.87           & 30.41            & \multicolumn{1}{c|}{36.20}             & 37.35            & 32.46            & 38.81             \\
              \multicolumn{1}{l|}{\textbf{SOV-STG-L}}                     & \multicolumn{1}{c|}{30}              & \multicolumn{1}{c|}{ResNet-101}      & 35.01           & 30.63            & \multicolumn{1}{c|}{36.32}             & 37.60            & 32.77            & 39.05             \\ 
              \multicolumn{1}{l|}{\textbf{SOV-STG-VLA-S}}                 & \multicolumn{1}{c|}{15}              & \multicolumn{1}{c|}{ResNet-50}       & \textbf{41.16}  & 39.48   & \multicolumn{1}{c|}{\textbf{41.67}}    & \textbf{43.81}   & \textbf{42.63}   & \textbf{44.17}  \\
              \rowcolor{Gray}
              \multicolumn{1}{l|}{\textbf{SOV-STG-Swin-L}}                & \multicolumn{1}{c|}{30}              & \multicolumn{1}{c|}{Swin-Large-22K}  & 43.35  & 42.25   & \multicolumn{1}{c|}{43.69}    & 45.53   & 43.62   & 46.11    \\ 
              \rowcolor{Gray}
              \multicolumn{1}{l|}{\textbf{SOV-STG-VLA-Swin-L}}            & \multicolumn{1}{c|}{15}              & \multicolumn{1}{c|}{Swin-Large-22K}  & \textbf{45.64}  & 44.35   & \multicolumn{1}{c|}{\textbf{46.03}}    & \textbf{48.22}   & 47.12   & \textbf{48.55}    \\ \hline
          \end{tabular}
      }
      \caption{
          Comparison to state-of-the-arts on the HICO-DET.
      }
      \label{tab:tab1}
      \resizebox{0.62\linewidth}{!}{
          \begin{tabular}{@{}cccccccccc@{}}
              \hline
              \multicolumn{1}{c|}{\multirow{2}{*}{\#}} & \multicolumn{1}{c}{\multirow{2}{*}{oDec}} & \multicolumn{1}{c}{\multirow{2}{*}{sDec}} & \multicolumn{1}{c}{\multirow{2}{*}{vDec}} & \multicolumn{1}{c}{\multirow{2}{*}{S-O Attn}} & \multicolumn{1}{c}{\multirow{2}{*}{STG}} & \multicolumn{1}{c|}{\multirow{2}{*}{VLA}} & \multicolumn{3}{c}{Default}\\
              \multicolumn{1}{c|}{}                    & \multicolumn{1}{c}{}                      & \multicolumn{1}{c}{}                      & \multicolumn{1}{c}{}                      & \multicolumn{1}{c}{}                          & \multicolumn{1}{c}{}                     & \multicolumn{1}{c|}{}                     & \multicolumn{1}{c}{\textit{Full}} & \multicolumn{1}{c}{\textit{Rare}} & \multicolumn{1}{c}{\textit{Non-Rare}} \\ \hline \hline
              \multicolumn{1}{c|}{(1)}                 & \multicolumn{1}{c}{\ding{51}}             & \multicolumn{1}{c}{}                      & \multicolumn{1}{c}{\ding{51}}             & \multicolumn{1}{c}{}                 & \multicolumn{1}{c}{\ding{51}}            & \multicolumn{1}{c|}{}                     & 32.68          & 28.21  & 34.02  \\ 
              \multicolumn{1}{c|}{(2)}                 & \multicolumn{1}{c}{\ding{51}}             & \multicolumn{1}{c}{\ding{51}}             & \multicolumn{1}{c}{}                      & \multicolumn{1}{c}{}                 & \multicolumn{1}{c}{\ding{51}}            & \multicolumn{1}{c|}{}                     & 32.35          & 27.64  & 33.63  \\
              \multicolumn{1}{c|}{(3)}                 & \multicolumn{1}{c}{\ding{51}}             & \multicolumn{1}{c}{}                      & \multicolumn{1}{c}{}                      & \multicolumn{1}{c}{}                 & \multicolumn{1}{c}{}                     & \multicolumn{1}{c|}{}                     & 30.14          & 22.82  & 32.32  \\
              \multicolumn{1}{c|}{(4)}                 & \multicolumn{1}{c}{\ding{51}}             & \multicolumn{1}{c}{\ding{51}}             & \multicolumn{1}{c}{}                      & \multicolumn{1}{c}{}                 & \multicolumn{1}{c}{}                     & \multicolumn{1}{c|}{}                     & 30.62          & 24.60  & 32.42  \\
              \multicolumn{1}{c|}{(5)}                 & \multicolumn{1}{c}{\ding{51}}             & \multicolumn{1}{c}{\ding{51}}             & \multicolumn{1}{c}{\ding{51}}             & \multicolumn{1}{c}{}                 & \multicolumn{1}{c}{}                     & \multicolumn{1}{c|}{}                     & 31.90          & 25.92  & 33.69  \\
              \multicolumn{1}{c|}{(6)}                 & \multicolumn{1}{c}{\ding{51}}             & \multicolumn{1}{c}{\ding{51}}             & \multicolumn{1}{c}{\ding{51}}             & \multicolumn{1}{c}{}                          & \multicolumn{1}{c}{\ding{51}}            & \multicolumn{1}{c|}{}                     & 33.01          & 27.83  & 34.55  \\
              \multicolumn{1}{c|}{(7)}                 & \multicolumn{1}{c}{\ding{51}}             & \multicolumn{1}{c}{\ding{51}}             & \multicolumn{1}{c}{\ding{51}}             & \multicolumn{1}{c}{\ding{51}}                 & \multicolumn{1}{c}{\ding{51}}            & \multicolumn{1}{c|}{}                     & 33.80 & 29.28 & 35.15 \\
              \multicolumn{1}{c|}{(8)}                 & \multicolumn{1}{c}{\ding{51}}             & \multicolumn{1}{c}{\ding{51}}             & \multicolumn{1}{c}{}             & \multicolumn{1}{c}{\ding{51}}                 & \multicolumn{1}{c}{\ding{51}}            & \multicolumn{1}{c|}{\ding{51}}            & 40.58 & 38.65 & 41.16 \\
              \multicolumn{1}{c|}{(9)}                & \multicolumn{1}{c}{\ding{51}}             & \multicolumn{1}{c}{\ding{51}}              & \multicolumn{1}{c}{\ding{51}}             & \multicolumn{1}{c}{\ding{51}}                 & \multicolumn{1}{c}{\ding{51}}            & \multicolumn{1}{c|}{\ding{51}}            & \textbf{41.16} & \textbf{39.48} & \textbf{41.67} \\ \hline
          \end{tabular}
      }
      \caption{Contributions of each module. The "oDec", "sDec", and "vDec" denote the object, subject, and verb decoder, respectively.}
      \label{tab:tab3}
  \end{minipage}
  \hfill
  \begin{minipage}{0.34\linewidth}
      \centering
      \resizebox{0.86\linewidth}{!}{
          \begin{tabular}{@{}cccc@{}}
              \hline
              \multicolumn{1}{c|}{Method}                                & \multicolumn{1}{c|}{Backbone} & \multicolumn{1}{c}{$AP_{role}^{S1}$} & \multicolumn{1}{c}{$AP_{role}^{S2}$}  \\ \hline \hline
              \multicolumn{1}{l|}{FGAHOI~\cite{ma2023fgahoi}}                 & \multicolumn{1}{c|}{ResNet-50}      & 59.0          & 59.3  \\
              \multicolumn{1}{l|}{RLIP-ParSe~\cite{yuan2022rlip}}            & \multicolumn{1}{c|}{ResNet-50}       & 61.9          & 64.2  \\
              \multicolumn{1}{l|}{MSTR~\cite{Kim_2022_CVPR}}                 & \multicolumn{1}{c|}{ResNet-50}       & 62.0          & 65.2  \\
              \multicolumn{1}{l|}{GEN-VLKT-S~\cite{liao2022gen}}             & \multicolumn{1}{c|}{ResNet-50}      & 62.4          & 64.5  \\
              \multicolumn{1}{l|}{GEN-VLKT-M~\cite{liao2022gen}}             & \multicolumn{1}{c|}{ResNet-101}      & 63.3          & 65.6  \\
              \multicolumn{1}{l|}{HOICLIP~\cite{ning2023hoiclip}}            & \multicolumn{1}{c|}{ResNet-50}       & 63.5          & 64.8  \\
              \multicolumn{1}{l|}{GEN-VLKT-L~\cite{liao2022gen}}             & \multicolumn{1}{c|}{ResNet-101}      & 63.6          & 65.9  \\
              \multicolumn{1}{l|}{CDN-L~\cite{zhang2021mining}}              & \multicolumn{1}{c|}{ResNet-101}      & 63.9          & 65.9  \\
              \multicolumn{1}{l|}{UniHOI-S~\cite{cao2024detecting}}          & \multicolumn{1}{c|}{ResNet-50}       & 65.6          & 68.3  \\ \hline
              \multicolumn{1}{l|}{\textbf{SOV-STG-S}}                        & \multicolumn{1}{c|}{ResNet-50}       & 61.2          & 62.5  \\
              \multicolumn{1}{l|}{\textbf{SOV-STG-M}}                        & \multicolumn{1}{c|}{ResNet-101}      & 63.7          & 65.2  \\
              \multicolumn{1}{l|}{\textbf{SOV-STG-L}}                        & \multicolumn{1}{c|}{ResNet-101}      & 63.9          & 65.4  \\
              \rowcolor{Gray}
              \multicolumn{1}{l|}{\textbf{SOV-STG-VLA-S}}                    & \multicolumn{1}{c|}{ResNet-50}       & 63.8          & 65.7  \\
              \hline
          \end{tabular}
      }
      \caption{Comparison on V-COCO.}
      \label{tab:tab2}
      \resizebox{\linewidth}{!}{
          \begin{tabular}{@{}ccccccc@{}}
              \hline
              \multicolumn{1}{c|}{\multirow{2}{*}{\#}} & \multicolumn{1}{c|}{Verb} & \multicolumn{1}{c|}{Q-Former} & \multicolumn{1}{c|}{\multirow{2}{*}{$\text{PE}(\bm{B}_v)$}} & \multicolumn{3}{c}{Default} \\ \cline{5-7}
              \multicolumn{1}{c|}{}    & \multicolumn{1}{c|}{Prediction} & \multicolumn{1}{c|}{Text Embed} & \multicolumn{1}{c|}{} & \multicolumn{1}{c}{\textit{Full}} & \multicolumn{1}{c}{\textit{Rare}} & \multicolumn{1}{c}{\textit{Non-Rare}} \\ \hline \hline
              \multicolumn{1}{c}{(1)}  & \multicolumn{1}{c}{\ding{51}}      & \multicolumn{1}{c}{\ding{51}}           & \multicolumn{1}{c|}{\ding{51}}     & 41.16 & 39.48 & 41.67 \\
              \multicolumn{1}{c}{(2)} & \multicolumn{1}{c}{\ding{51}} & \multicolumn{1}{c}{\ding{51}} & \multicolumn{1}{c|}{} & 40.47 & 37.85  & 41.25 \\
              \multicolumn{1}{c}{(3)} & \multicolumn{1}{c}{\ding{51}} & \multicolumn{1}{c}{} & \multicolumn{1}{c|}{\ding{51}} & 40.17 & 37.25  & 41.04 \\
              \multicolumn{1}{c}{(4)} & \multicolumn{1}{c}{} & \multicolumn{1}{c}{\ding{51}} & \multicolumn{1}{c|}{\ding{51}} & 40.21 & 38.01 & 40.86 \\
              \multicolumn{1}{c}{(5)} & \multicolumn{1}{c}{} & \multicolumn{1}{c}{} & \multicolumn{1}{c|}{\ding{51}} & 39.78 & 35.51  & 41.05 \\ \hline
          \end{tabular}
      }
      \caption{Ablation studies of VLA.}
      \label{tab:tab_v-hoi}
      \resizebox{0.71\linewidth}{!}{
          \begin{tabular}{@{}ccccc@{}}
              \hline
              \multicolumn{1}{c|}{\multirow{2}{*}{\#}} & \multicolumn{1}{c|}{\multirow{2}{*}{Verb Box}} & \multicolumn{3}{c}{Default}  \\
              \multicolumn{1}{c|}{}                    & \multicolumn{1}{c|}{}                          & \textit{Full} & \textit{Rare} & \textit{Non-Rare} \\ \hline \hline
              \multicolumn{1}{c|}{(1)}                 & \multicolumn{1}{c|}{Object Box}                & 33.16  & 27.21  & 34.94    \\
              \multicolumn{1}{c|}{(2)}                 & \multicolumn{1}{c|}{Subject Box}               & 32.78  & 28.01  & 34.21    \\
              \multicolumn{1}{c|}{(3)}                 & \multicolumn{1}{c|}{MBR}                       & 33.44  & 27.84  & 35.11    \\ 
              \multicolumn{1}{c|}{(4)}                 & \multicolumn{1}{c|}{SMBR}                      & 33.41  & 28.22  & 34.97    \\ 
              \multicolumn{1}{c|}{(5)}                 & \multicolumn{1}{c|}{ASMBR}                     & \textbf{33.80}  & \textbf{29.28}  & \textbf{35.15} \\ \hline
          \end{tabular}
      }
      \caption{Different designs for the verb box. The "SMBR" indicates the Shifted MBR.}
      \label{tab:tab4}
      \resizebox{0.81\linewidth}{!}{
          \begin{tabular}{@{}ccccccc@{}}
              \hline
              \multicolumn{4}{c|}{Denoising Strategies}                                      & \multicolumn{3}{c}{Default} \\ \cline{1-4}
              \multicolumn{1}{c|}{\#}    & \multicolumn{1}{c}{Box} & \multicolumn{1}{c}{Obj} & \multicolumn{1}{c|}{Verb} & \multicolumn{1}{c}{\textit{Full}} & \multicolumn{1}{c}{\textit{Rare}} & \multicolumn{1}{c}{\textit{Non-Rare}} \\ \hline \hline
              \multicolumn{1}{c}{(1)}    & \multicolumn{1}{c}{}              & \multicolumn{1}{c}{}                      & \multicolumn{1}{c|}{}          & 32.99   & 28.28  & 34.40  \\
              \multicolumn{1}{c}{(2)}    & \multicolumn{1}{c}{\ding{51}}     & \multicolumn{1}{c}{}                      & \multicolumn{1}{c|}{}          & 33.27   & 29.07  & 34.53  \\
              \multicolumn{1}{c}{(3)}    & \multicolumn{1}{c}{\ding{51}}     & \multicolumn{1}{c}{}                      & \multicolumn{1}{c|}{\ding{51}} & 33.28   & 28.57  & 34.69  \\
              \multicolumn{1}{c}{(4)}    & \multicolumn{1}{c}{}              & \multicolumn{1}{c}{\ding{51}}             & \multicolumn{1}{c|}{\ding{51}} & 33.39   & 28.82  & 34.76  \\
              \multicolumn{1}{c}{(5)}    & \multicolumn{1}{c}{\ding{51}}     & \multicolumn{1}{c}{\ding{51}}             & \multicolumn{1}{c|}{}          & 33.51   & 29.05  & 34.84  \\
              \multicolumn{1}{c}{(6)}    & \multicolumn{1}{c}{\ding{51}}     & \multicolumn{1}{c}{\ding{51}}             & \multicolumn{1}{c|}{\ding{51}} & \textbf{33.80}   & \textbf{29.28}  & \textbf{35.15}  \\ \hline      
          \end{tabular}
      }
      \caption{Ablation studies of denoising strategies. The symbol of \ding{51} means adding noise to the ground-truth.}
      \label{tab:tab6}
  \end{minipage}
\end{table*}

\subsection{Experiments}
We evaluate our proposed SOV-STG on the HICO-DET~\cite{chao2018learning} and V-COCO~\cite{gupta2015visual} datasets to compare with other SOTA methods and conduct extensive ablation studies to analyze the contributions of each component and show the effectiveness of our proposed method.

\subsection{Experimental Settings}
\noindent{\textbf{Dataset and Metric.}}\quad The HICO-DET~\cite{chao2018learning} dataset contains 38,118 images for training and 9,658 images for the test.
The 117 verb classes and 80 object classes in HICO-DET form 600 HOI classes.
According to the number of HOI instances appearing in the dataset, the HOI classes are divided into three categories: \textit{Full}, \textit{Rare}, and \textit{Non-Rare}.
Moreover, considering HOI instances including or not including the unknown objects, the evaluation of HICO-DET is divided into two settings: Default and Known Object.
The V-COCO~\cite{gupta2015visual} dataset contains 5,400 images for training and 4,946 images for the test.
In V-COCO, 80 object classes and 29 verb classes are annotated, and two scenarios are considered: scenario 1 with 29 verb classes and scenario 2 with 25 verb classes.
We follow the standard evaluation~\cite{chao2018learning} and report the mAP scores.

\noindent{\textbf{Implementation Details.}}\quad We use the DAB-Deformable-DETR trained on COCO~\cite{lin2014microsoft} to initialize the weight of the feature extractor, the subject decoder, and the object decoder.
The feature extractor consists of a ResNet-50~\cite{he2016deep} backbone and a 6-layer deformable transformer encoder.
We implement three variants of SOV-STG, which are denoted as \textbf{SOV-STG-S} with ResNet-50 and 3-layer decoders, \textbf{SOV-STG-M} with ResNet-101 and 3-layer decoders, and \textbf{SOV-STG-L} with ResNet-101 and 6-layer decoders.
The hidden dimension of the transformer is $D=256$, and the number of the query $N_q$ is set to 64 for HICO-DET and 100 for V-COCO.
For the DN part, $2N_p=6$ groups of noised labels are generated for each ground-truth HOI instance.
The same as previous works~\cite{ning2023hoiclip,mao2024clip4hoi,cao2024detecting}, we use ResNet-50 as the backbone and add VLA to our SOV-STG-S to form \textbf{SOV-STG-VLA-S}.
In addition, we also use Swin-Transformer~\cite{liu2021swin} as the backbone to achieve the best performance.
The ViT-32/B variant is used as the visual encoder in our VLA.
The vision advisor decoder has the same number of layers as SOV decoders.
We train SOV-STG with the AdamW optimizer~\cite{loshchilov2018decoupled} with a learning rate of 2e-4 (except for the backbone, which is 1e-5 for HICO-DET, 2e-6 for V-COCO) and a weight decay of 1e-4.
The batch size is set to 32 (8 NVIDIA A6000 GPUs, 4 images per GPU), and the training epochs are 30 (the learning rate drops at the 20th epoch).
We train SOV-STG-VLA with half of the learning rate, batch size, and training epochs.

\subsection{Comparison to State-of-the-Arts}
In \cref{tab:tab1}, we compare SOV-STG and SOV-STG-VLA with the recent SOTA methods on the HICO-DET dataset.
Our SOV-STG-S with ResNet-50 backbone achieves 33.80 mAP on the \textit{Full} category of the Default setting.
Compared with the transformer-based one-stage methods, QAHOI and MSTR, which are based on the reference point,
SOV-STG benefits from the anchor box priors and label priors and achieves 7.62 (29.11\%) and 2.63 (8.44\%) mAP improvements, respectively.
Note that, without any extra language prior knowledge~\cite{radford2021learning}, SOV-STG-M outperforms GEN-VLKT-M by 0.26\% in one-third of the training epochs.
Since our SOV-STG explicitly makes full use of the ground-truth information, compared with DOQ, which also uses ground-truth to guide the training, SOV-STG-S achieves 1.05\% mAP improvement with less than half of the training epochs of DOQ.
Especially, compared with UniHOI-S and GEN-VLKT-S, UniHOI gains 18.70\% mAP with the use of VLM and additional modules.
Our VLA effectively connects the verb prediction part of SOV-STG and VLM, as a result, SOV-STG-VLA-S achieves 41.16 mAP and improves the performance by 21.8\% compared with the SOV-STG-S, which is higher than the improvement of UniHOI-S, and our training epochs is one-sixth of UniHOI-S.
Moreover, our SOV-STG-VLA-Swin-L achieves 45.64 mAP also with only 15 training epochs.
In \cref{tab:tab2}, on the V-COCO dataset, our SOV-STG-VLA-S improves SOV-STG-S by 4.25\% and 5.12\% on the scenario 1 and 2, respectively.

\subsection{Ablation Study}

We conduct all the ablation experiments on the HICO-DET dataset with the SOV-STG-S model.

\noindent{\textbf{Contributions of proposed modules.}}\quad SOV-STG is composed of flexible decoding architecture and training strategies.
To clarify the contributions of each proposed module, in \cref{tab:tab3}, we remove the proposed modules one by one and conduct ablation studies on the HICO-DET dataset.
The row of (5) indicates the experiment removing the STG strategy and the S-O attention module is degraded to a sum fusion module.
From the result, compared with SOV-STG in (7), the STG strategy and S-O attention improve the performance by 5.96\% on the \textit{Full} category.
Next, in (4), we remove the verb decoder in (5).
As a result, comparing (4) with (5), without the verb decoder, the performance drops by 4.01\%.
Then, in (3), we remove the subject decoder and the sum fusion module and update both the subject and object boxes by the object decoder.
Without balancing the decoding burden of the detection, compared with (4), the performance drops by 1.57\%.
Furthermore, in (1) and (2), we conduct drop-one-out experiments on the subject and verb decoder, respectively.
Compared with (1) and (2), the model without the verb decoder is worse than the model without the subject decoder, which indicates that the verb decoder plays a more critical role.
With the MSACrossAttn in our vision advisor decoder, our vision advisor decoder can extract specific region features as the verb decoder, therefore, in (8), we remove the verb decoder of SOV-STG-VLA-S, and the model still performs well.

\noindent{\textbf{Vision Language Advisor.}}\quad Our VLA connects the verb and HOI label prediction both in the visual and language space.
To verify the effectiveness of the VLA, we conduct ablation studies in \cref{tab:tab_v-hoi}.
In row (2), we remove the positional embeddings from the vision advisor decoder, and from the result, the performance drops by 4.13\% on the \textit{Full} category.
In row (3), we train the HOI class head from scratch without initializing the weight from the text embeddings.
From the result, without the guidance of Q-Former in language, the performance on the \textit{Rare} category drops by 5.65\%.
In row (4), we investigate the effect of the connection of verb and HOI prediction by removing the verb prediction part from VLA.
Specifically, the verb embeddings $E_{v'}$ from the verb decoder are directly added to the refined verb embeddings $E_{va}$ from the vision advisor decoder and fed into the HOI class head.
As a result, the performance drops by 2.31\% on the \textit{Full} category and 3.72\% on the \textit{Rare} category compared with the full VLA in (1).
Then, in row (5), we remove the verb prediction and train the HOI class head from scratch, and the performance on the \textit{Rare} category drops by 10.06\% compared with row (1).

\noindent{\textbf{Denoising Strategy.}}\quad In \cref{tab:tab6}, we investigate the denoising strategies of three parts of the targets, i.e., the box coordinates, the object labels, and the verb labels.
The result of (6) indicates the result of SOV-STG-S.
In (1), we set the noise rate of box coordinates to $\delta_b=0$, the object label flipping rate to $\eta_o=0$, and the verb label flipping rate to $\eta_v=0$, thus, the ground-truth box coordinates, object labels, and verb labels are directly fed into the model without any noise.
From the result, the accuracy drops by 2.40\% compared with the full denoising training in (6).
In (3), (4), and (5), we conduct drop-one-out experiments, and the results show that each part of the denoising strategy is effective.
For the results between (2) and (3), and (4) and (3), the verb denoising increases the performance when it is used with the object denoising.

\noindent{\textbf{Formulations of the verb box.}}\quad To verify the effectiveness of ASMBR, we use the verb box degraded from the ASMBR to conduct ablation studies, and the results are shown in \cref{tab:tab4}.
From the results of (3) to (5), the adaptive and shift operations for the MBR promote the performance of the verb box, by 1.08\% on the \textit{Full} category and 5.17\% on the \textit{Rare} category.
Furthermore, in (1) and (2), we directly use the object or subject box as the verb box, and the results show that the object box is better for non-rare class recognition, while the subject box is better for rare class recognition.

\section{Conclusion}
In this paper, we propose a novel one-stage HOID framework, SOV for target-specific decoding, and a specific target guided denoising strategy, STG, for efficient training.
We introduce a new format to represent the interaction region in a verb box and improve the verb prediction.
Our SOV-STG shows the effectiveness of splitting the HOI decoding with each element of the HOI triplet.
Based on SOV-STG, we further propose VLA, which integrates VLM to enhance the interaction representation.
Our vision advisor focuses on the alignment between the VLM's prior knowledge and the verb embeddings with the guidance of spatial information.
Our V-HOI Bridge connects the visual and language information for verb and HOI label prediction.
With the well-designed architecture and training strategy, our framework achieves state-of-the-art performance with fast convergence.
In the future, we will explore the potential of the proposed framework in other vision-language tasks and further improve the performance of the model.

{\small
\bibliographystyle{ieee_fullname}
\bibliography{egbib}
}

\end{document}